\documentclass[letterpaper]{article} 
\usepackage{aaai24}  
\usepackage{times}  
\usepackage{helvet}  
\usepackage{courier}  
\usepackage[hyphens]{url}  
\usepackage{graphicx} 
\usepackage{natbib}  
\usepackage{caption} 
\usepackage{algorithm}
\usepackage{algorithmic}

\usepackage{amsmath}
\usepackage{amsfonts}
\usepackage{color}

\usepackage{newfloat}
\usepackage{listings}
\DeclareCaptionStyle{ruled}{labelfont=normalfont,labelsep=colon,strut=off} 
\floatstyle{ruled}
\newfloat{listing}{tb}{lst}{}
\floatname{listing}{Listing}

\title{Fooling SHAP with Output Shuffling Attacks}
\author{
    Jun Yuan and Aritra Dasgupta
}
\affiliations{
    New Jersey Institute of Technology (NJIT)\\
    Department of Data Science

    jy448@njit.edu, aritra.dasgupta@njit.edu
}

\usepackage{bibentry}
\usepackage{xcolor}

\begin{document}

\maketitle

\begin{abstract}
    Explainable AI~(XAI) methods such as SHAP can help discover feature attributions in black-box models. If the method reveals a significant attribution from a ``protected feature'' (e.g., gender, race) on the model output, the model is considered unfair. However, adversarial attacks can subvert the detection of XAI methods. Previous approaches to constructing such an adversarial model require access to underlying data distribution, which may not be possible in many practical scenarios. We relax this constraint and propose a novel family of attacks, called shuffling attacks, that are data-agnostic. The proposed attack strategies can adapt any trained machine learning model to fool Shapley value-based explanations. 
    
    We prove that Shapley values cannot detect shuffling attacks. However, algorithms that estimate Shapley values, such as linear SHAP and SHAP, can detect these attacks with varying degrees of effectiveness. We demonstrate the efficacy of the attack strategies by comparing the performance of linear SHAP and SHAP using real-world datasets. 
\end{abstract}

\section{Introduction}

Explainable AI~(XAI) methods are increasingly being used to detect unfairness in black-box machine learning models~\cite{arrieta2020explainable,das2020opportunities}. For example, suppose the XAI method detects that ``protected features'' such as gender or race significantly contribute to a model's prediction. In that case, it may indicate that the model is unfair. 

Existing work shows adversarial classifiers can be constructed via scaffolding~\cite{slack2020fooling}, which can fool explanations generated by LIME~\cite{ribeiro2016should} and SHAP~\cite{lundberg2017unified}. Specifically, LIME and SHAP cannot detect that the classifier is making decisions heavily influenced by the ``protected feature''. However, a scaffolding procedure assumes we have access to the underlying dataset, which can be used to train an additional classification model. 
Such an assumption might not hold good in real-world scenarios where data accessibility is restricted.
For example, while the training data for a machine learning model remains confidential, the types of input data (including both protected and unprotected attributes) can often be deduced from the model's user instruction manual. This method of releasing models is frequently seen in large language models like Meta's Llama series.

In this work, we focus on scoring functions that assign probability or scores to data items, where the scores can either be used to assign items to different groups~(in the case of a classifier) or put them in a particular order~(in the case of an algorithmic ranker).

Scoring functions represent a more fine-grained model output than the model predictions themselves.
We can turn a scoring function into a classifier by setting thresholds on the score output~(e.g., if the score is above a threshold, it belongs to class A, otherwise class B.).
Also, when the input of the scoring function is a group of items rather than a single item, the resulting score vector can be used to generate a ranking~(indicating the relative preference for the items). Hence, the scoring function can also be considered an intermediate component of an ``algorithmic ranker''~\cite{yuan2023trivea}.

Our contributions can be summarized as follows: 1) We propose and algorithmically demonstrate a number of model output shuffling attacks,
2) Theoretically, we prove that Shapely values cannot detect our proposed attacks.
3) Experimentally, we show the impact of the attack on various 
detection 
settings using multiple real-world datasets.

\section{Related Work}
The goal in this work is to modify a base function $f$ without accessing the training data of $f$~(and even the internal structure of $f$) and achieve adversarial function $f'$ that modifies the outcome of input $\textbf{x} \in \textbf{X}$. However, an XAI method $g()$, which requires $f$ as one of the inputs, considers $f'$ to behave similarly to $f$. In particular, the protected features' attribution is smaller than a certain threshold. 
Based on a recent survey paper~\cite{baniecki2023adversarial} that provides a unified notation and taxonomy of adversarial attacks on XAI, our proposed method can be formulated as:
\begin{equation}
    f \rightarrow f' \Rightarrow 
    \begin{cases} 
        \forall \textbf{x} \in \textbf{X} & g(f, \textbf{x}) \approx g(f', \textbf{x}) \\
        \exists \textbf{x} \in \textbf{X} & f(\textbf{x}) \neq f'(\textbf{x}) \\
    \end{cases}
    \label{eqn:advXAI_type}
\end{equation}

The survey describes other adversarial works in a similar notation as 
Formula~(Eq.\ref{eqn:advXAI_type}). 
However, no existing work can be described as formula~(Eq.\ref{eqn:advXAI_type}) exactly. 
Hence, our method is unique from other works.
The survey includes one work~\cite{noppel2023disguising} that is most similar to our method~(Eq.\ref{eqn:advXAI_type}) but relies on poisoning data used in training~(e.g., in a scenario of outsourced training) to trigger wrong outputs for certain input.
For the attack methods that construct models without changing data, 
\cite{dimanov2020you} attempt to learn a new model that performs similarly to the original but has a much lower target~(e.g., gender, race) feature attribution. The resulting new model's behavior is unclear whether improving or degrading the fairness of the original. They use common fairness metrics to conduct post-hoc evaluations to compare the fairness shifts between original and new models.
Our method, however, constructs specific unfair behavior of the new model without any training. 
This allows the attack to be performed not only by the data owners themselves. Instead, the attack can be performed by model distributors or brokers.
Other works~(\cite{slack2020fooling}, ~\cite{laberge2023fool}) attempt to deceive explanation methods using deception mechanisms that exploit the SHAP's permutation strategy on data or data distribution.
Our method does not require accessing data or data distribution, training a new model, or changing the data.

\section{Attack Strategies}
Existing works on adversarial attacks on XAI methods assume the adversary action is executed by the model developers~\cite{slack2020fooling} or the data owners~\cite{aivodji2022fooling}.
We consider the case in which the adversary action is executed by the model distributors or the model brokers.
Model users may only access the model via the API provided by the model brokers.
We assume model brokers~(e.g., companies hosting their models using cloud services) do not access or change the model architecture or the underlying training data but may gain input from the model users.

In this section, we introduce the Shapley values. Then, we provide the intuition of the shuffling attack and three examples of it. 
We discuss the connections among the attacks, the relative strength of the attack, and ways to modify the attack to bypass detection.
\subsection{Shapley values}
Shapley values originated from cooperative game theory. 
Intuitively, Shapley values attribute payouts to a game's players.
Let $\mathbf{X}$ contain $d$ random variables corresponding to features in set $\mathcal{D}$.
A coalition $\mathcal{S}$ is a subset of the total feature set $\mathcal{D}$.
The $\mathbf{X}_{\mathcal{S}}$ and  $\mathbf{X}_{\mathcal{\bar{S}}}$ denote in-coalition and out-of-coalition feature variables.
Considering different combinations~(i.e. coalition) of a team, the average change in the output of the model when a player~(i.e. feature) joins the team is its payout or Shapley value:
\begin{equation}
    \phi_j = \sum_{\mathcal{S} \subseteq \mathcal{D} \setminus \{j\}} \frac{
        |\mathcal{S}|! (|\mathcal{D}| - |\mathcal{S}| - 1)! 
    }{ 
        |\mathcal{D}|!
    }
    (v(\mathcal{S} \cup \{j\}) - v(\mathcal{S})  ).
    \label{eqn:shapley_value}
\end{equation}

Both $\mathcal{S}$ and $\mathcal{S} \cup \{j\}$ are subsets of the total feature set $\mathcal{D}$.
The difference between $v(S)$ and $v(S \cup \{j\})$ is the payout of individual feature $j$ in one coalition.

Shapley value of feature $j$ is a weighted sum of payout differences between coalition $S$ and $\mathcal{S} \cup \{j\}$, for weight is the frequency of such coalition $\mathcal{S}$ in random selection. 
A payout is calculated by the value function $v(\cdot)$ on a certain coalition.
Commonly defined value function may be the expectation with respect to a distribution $\mathcal{R}$ that can be inferred from data or prior knowledge. 

For an observed $\mathbf{x}_{\mathcal{S}}$ , the payout of coalition $\mathcal{S}$ is

\begin{equation}
    v(\mathcal{S}) = \mathbb{E}_{\mathcal{R}}[f(\mathbf{x}_{\mathcal{S}}, \mathbf{X}_{\mathcal{\bar{S}}} )].
    \label{eqn:value_func}
\end{equation}

\subsection{Adversarial shuffling}
Shuffling attacks exploit the order-agnostic nature of the expectation calculation for value function $v$.
For $N$ samples of $y_i$ that
\begin{equation}
    y_i = f((\mathbf{x}_{\mathcal{S}}, \mathbf{X}_{\mathcal{\bar{S}}})_{i}), i= 1,2,\cdots, N.
    \label{eqn:scoring_func}
\end{equation}
The vector of $y_i$ is $\textbf{y}$. We rewrite the equation~(\ref{eqn:value_func}) as: 
\begin{equation}
    v(\mathcal{S}) = \frac{1}{N}\sum_i y_i = \frac{1}{N} \textbf{1}^T \textbf{y} .
\end{equation}

Let $\text{shuffle}_{\{j\}}$ 
denote an operator that transforms a vector 
$\textbf{y}$
to a permutation 
$\textbf{y}'$ 
and depends on feature $j$ or the vector $\textbf{X}_j$ only. 
In other word, feature $j$ determines the $\text{shuffle}_{\{j\}}$  operator.
For example, 
between the original order of the output $\textbf{y} = [y_1, y_2, \cdots, y_n]$ and the permuted order of the output $\textbf{y}' = [y'_1, y'_2, \cdots, y'_n]$, there is a one-to-one mapping between each element of $\textbf{y}$ and $\textbf{y}'$. 

The mapping is defined based on values in feature $j$, $X_j = [x_{1j}, x_{2j}, \cdots, x_{nj}]$. When all the elements of $X_j$ are the same, the shuffle operator is not executed. When the length of $\textbf{y}$ is 1, $n=1$, the shuffle operator is not executed.
For simplicity, we assume feature $j$ does not affect the calculation in equation~(\ref{eqn:scoring_func}).

Assuming we want to calculate the attribution of feature $j$,
since shuffle operator neither affects the total sum of $\textbf{y}$ nor changes the expectation, we get
\begin{equation}
    v(\mathcal{S} \cup \{j\}) - v(\mathcal{S}) = \frac{1}{N} \textbf{1}^T \textbf{y}'
    - \frac{1}{N} \textbf{1}^T \textbf{y} = 0 
    \label{eqn:payout_zero}
\end{equation}

We plug equation~(\ref{eqn:payout_zero}) in ~(\ref{eqn:shapley_value}), and get attribution of feature $j$, $\phi_j$, is always zero.
Although neglected by Shapley values, the shuffling effect on $\textbf{y}$ can reflect unfair treatment between groups. 
For example, it may describe an unfair allocation of bonuses, such as a manager swapping bonuses between male and female employees, although the females contributed more. 
Since the gender's attribution is non-detectable based on Shapley values, the manager gets away.

Theoretically, we may state that shuffling will never be detected by Shapley values. However, common implementations of Shapley values, such as linearSHAP and SHAP, perform certain estimations that can detect non-zero attributions from shuffling features. 

We experimented with different shuffling strategies on real-world datasets and different SHAP implementations.
In general, Shapley estimators or implementations of Shapley values take inputs of a given instance ID $i$, a background data $\mathbf{X}_{\mathcal{D}}$, and return a row vector $\Phi_{i}$ for the instance $i$.

In this work, we focus on the manipulation of the behavior of $f$.
For example, the parameters~(or code) of the function $f$ can be random in each while-loop, or adapted to data $\textbf{X}_{\mathcal{D'}}$ totally or partially.

\subsection{Shuffling Attack}

We introduce three types of shuffling attacks in pseudocode.
We denote the 
protected features $\mathcal{P}$~(e.g. race, gender), and non-protected features $\mathcal{D} \setminus \mathcal{P}$.
We define function $f$ and function $f'$, so an XAI method (e.g., SHAP) that can explain $f$ can also explain  $f'$. 

\begin{equation}
    \mathbf{y} = f(\mathbf{X}_{\mathcal{D}}) = f(\mathbf{X}_{\mathcal{D} \setminus \mathcal{P}}, \mathbf{X}_{{\mathcal{P}}} = \mathbf{0}) 
\end{equation}
\begin{equation}
    \mathbf{y}' = f'(\mathbf{X}_{\mathcal{D}}) = h_{\mathbf{X}_{\mathcal{P}}}( f(\mathbf{X}_{\mathcal{D} \setminus \mathcal{P}}))
\end{equation}

$f$ is a scoring function that does not use $\mathbf{X}_\mathcal{P}$. $f$ can be a simple linear model or a complex neural network that takes input data and returns a score or vector of scores.

Function $f$ may be used to assign probability or scores to data items. Scores can be used to i) assign items to different groups~(e.g., a classifier) by setting thresholds or ii) order the items~(e.g., an algorithmic ranker) using the score vector.

Function $f'$ is a wrapper of $f$, and contains a shuffling function $h$.
Both the $f$ and $f'$ functions take the entire data as input. However, the attack integrated in function $f'$ does not require data other than access to the protected features.

Function $h$ first needs to determine the superior outcomes. 
For example, if $f$ outputs the admission rate, a higher value is superior.

With such information, it can sort the vector $\mathbf{y}$.
The second step is determining the shuffling strategy to produce the  $\mathbf{y}'$. 
Due to the complexity and various choices involved in constructing function $f'$, we first provide a pseudocode template~(Algorithm \ref{alg:template}).

Line 5 calls an Attack function.
We further provide three Attack functions~(Algorithms \ref{alg:Dominance}, \ref{alg:Mixing}, \ref{alg:swap}).
To ensure clarity and consistency of the codes, we define all attacks based on a two-class gender bias scenario.
Each attack describes a decision-maker who prefers the male group over the female group. 

Our definition and codes can be generalized beyond the two-class gender scenario to race, age, or other protected features and their intersections.

\begin{algorithm}[tb]
    \caption{$f'$ algorithm template}
    \label{alg:template}
    \textbf{Input}: Dataset $\textbf{X}_{\mathcal{D}}$, function $f$\\
    \textbf{Parameter}: protected features $\mathcal{P}$, higher/lower $f$ output is superior\\
    \textbf{Output}: Unfair score vector $\textbf{y}'$
    \begin{algorithmic}[1]
        \STATE $\textbf{y} \gets f(\textbf{X}_{\mathcal{D}})$
        \STATE $N \gets \text{length of } \textbf{y}$ 
        \STATE $\textbf{id} \gets [1,2,3,...,N]$ 
        \STATE $\textbf{p} \gets \textbf{X}_\mathcal{P}$ 
        \STATE $[\textbf{id}, \textbf{p}, \textbf{y}] \gets \text{Sort}([\textbf{id}, \textbf{p}, \textbf{y}])  \text{ so the superior }  \textbf{y}_i  \text{ is on top}$  
        \STATE $\textbf{y}' \gets \text{Attack}([\textbf{id}, \textbf{p}, \textbf{y}])$
        \STATE \textbf{return} $\textbf{y}'$
    \end{algorithmic}
\end{algorithm}

\begin{algorithm}[tb]
    \caption{Attack: Dominance algorithm}
    \label{alg:Dominance}
    \textbf{Input}: [\textbf{id}, \textbf{p}, \textbf{y}]\\
    \textbf{Output}: Unfair score vector $\textbf{y}$
    \begin{algorithmic}[1] 
    \STATE $\textbf{y} \gets [\textbf{id}, \textbf{p}, \textbf{y}]$
    \STATE $[\textbf{id}, \textbf{p}] \gets [\textbf{id}, \textbf{p}, \textbf{y}]$
    \STATE $[\textbf{id}, \textbf{p}] \gets \text{Sort}([\textbf{id}, \textbf{p}] )  \text{ so the \textit{males}}  \text{ are on top}$ 
    \STATE $ [\textbf{id}, \textbf{p}, \textbf{y}] \gets \text{Concat}([\textbf{id}, \textbf{p}], \textbf{y})$
    \STATE $[\textbf{id}, \textbf{y}] \gets [\textbf{id}, \textbf{p}, \textbf{y}]$ 
    \STATE $\textbf{y} \gets \text{Sort}([\textbf{id}, \textbf{y}]) \text{ by } \textbf{id} $
    \STATE \textbf{return} \textbf{y} 
    \end{algorithmic}
\end{algorithm}

\begin{algorithm}[tb]
    \caption{Attack: Mixing algorithm}
    \label{alg:Mixing}
    \textbf{Input}: [\textbf{id}, \textbf{p}, \textbf{y}]\\
    \textbf{Output}: Unfair score vector $\textbf{y}$
    \begin{algorithmic}[1] 
    \STATE  $\textbf{y} \gets [\textbf{id}, \textbf{p}, \textbf{y}]$
    \STATE $\textbf{id}_{\textit{male}}$, $\textbf{id}_{\textit{female}} \gets [\textbf{id}, \textbf{p}, \textbf{y}] $ 
    \STATE \textbf{newId} = []
    \WHILE{$\textbf{id}_{\textit{male}} \text{ not empty and } \textbf{id}_{\textit{female}} \text{ not empty}$}
        
        \IF{unfairCoin($p$)}
            \STATE \textbf{newId} appends pop($\textbf{id}_{\textit{male}}$)
            \STATE pop($\textbf{y}_{\textit{male}}$)
        \ELSIF{$\textbf{y}_{\textit{female}}[0] \leq \textbf{y}_{\textit{male}}
        [0]$}
            \STATE \textbf{newId} appends pop($\textbf{id}_{\textit{male}}$)
            \STATE pop($\textbf{y}_{\textit{male}}$)
            
        \ELSE
            \STATE \textbf{newId} appends pop($\textbf{id}_{\textit{female}})$
            \STATE pop($\textbf{y}_{\textit{female}}$)
        \ENDIF 
    \ENDWHILE
    \IF {$\textbf{id}_{\textit{male}} \text{ not empty}$}
        \STATE \textbf{newId} extends $\textbf{id}_{\textit{male}}$
    \ELSE 
        \STATE \textbf{newId} extends $\textbf{id}_{\textit{female}}$
    \ENDIF
    \STATE $\textbf{y} \gets \text{Sort}([\textbf{newId}, \textbf{y}]) \text{ by } \textbf{newId} $
    \STATE \textbf{return} \textbf{y} 
    \end{algorithmic}
\end{algorithm}

\begin{algorithm}[tb]
    \caption{Attack: Swapping algorithm}
    \label{alg:swap}
    \textbf{Input}: [\textbf{id}, \textbf{p}, \textbf{y}]\\
    \textbf{Output}: Unfair score vector $\textbf{y}$
    \begin{algorithmic}[1] 
    \STATE $N \gets \text{length of vector } \textbf{y}$
    \FOR{ $i = 0$ to $N-2$}
    \IF{$\textbf{p}[i] \text{ is \textit{female} and } \textbf{p}[i+1] \text{ is \textit{male}}$}
        \STATE swap($\textbf{p}[i],\textbf{p}[i+1]$)
        \STATE swap($\textbf{id}[i],\textbf{id}[i+1]$)
    \ENDIF
    \ENDFOR
    \STATE $\textbf{y} \gets \text{Sort}([\textbf{id}, \textbf{y}]) \text{ by } \textbf{id} $
    \STATE \textbf{return} \textbf{y} 
    \end{algorithmic}
\end{algorithm}

\par \noindent \textbf{Dominance.}
Algorithm~\ref{alg:Dominance}
describes a \textit{biased decision-maker}, who see a sorted list of \textit{female} and \textit{male} candidates, gives all the \textit{male} candidates the high scores and \textit{female} low scores~(i.e. the last \textit{male} is ranked one position higher than the first \textit{female}), giving \textit{male} candidates an unfair advantage. 

The output of this algorithm is a list of scores. But if we impose a threshold to convert the scores to a two-class case~(e.g. approve and disapprove), it resembles the unfair classifier in the prior work~\cite{slack2020fooling}.
\par \noindent \textbf{Mixing.}
Algorithm~\ref{alg:Mixing}
describes a \textit{biased decision-maker}, who see a sorted list of \textit{female} and \textit{male} candidates, gives \textit{male} candidates a higher chance~(with probability $p$) of getting higher scores than \textit{female}.
Given the current highest score, the \textit{biased decision-maker} tosses a coin with probability $p$ that ``head" leads to a \textit{male} receiving the score. 
If the tossed outcome is ``tail", the \textit{biased decision-maker} compares the original score difference between the two candidates, the one with a higher score will be given the new score.
A $p$ larger than $0.5$ gives \textit{male} an unfair advantage.
Dominance attack is a special case of Mixing attack when $p$ is $1$. 
The Mixing algorithm is inspired by~\cite{yangMeasuringFairnessRanked2017}.

\par \noindent \textbf{Swapping.}
Algorithm \ref{alg:swap}
describes a \textit{biased decision-maker}, who see a \textit{female} candidate is scored higher than a \textit{male} candidate~(i.e. a \textit{female} is ranked one position higher than a \textit{male}), swaps the two candidates' scores~(and consequently their rank positions are swapped), giving \textit{male} candidates an unfair advantage. 

The Swapping attack may be triggered zero or many times. 
If all \textit{male}'s scores are larger than \textit{female}'s, no Swapping occurs. If all candidates are \textit{female} or \textit{male}, no Swapping occurs. 
But if there is only one \textit{female} in the data who scored highest~(i.e., at the top-1 rank position) among other \textit{male} candidates, the \textit{female} candidate's score will be swapped repeatedly and ends up at the bottom of the ranking with the lowest score. In such a case, the Swapping attack is equivalent to a Dominance attack.

The Dominance attack can be described as a more complex Swapping attack in general by adding one additional for-loop
and move the line 2 to 7 of Algorithm~\ref{alg:swap} inside the for-loop. 
The for-loop condition is $j=N-2 \text{ to } 0$ 
We rewrite the inner for-loop condition as $i=j \text{ to } N-2$.
In this way, the algorithm finds the \textit{female} candidates from the bottom to the top of the score list and swaps them repeatedly to the bottom of the list.

In other words, the Swapping attack is a special case of Dominance attack when the outer for-loop only executes once at $j = 0$. 
The starting point of $j$ can be determined by a quantile of a given score list, meaning $j=N$ corresponds to quantile 1, and $j=0$ to quantile 0.

\subsection{Techniques to modify the attacks}

We describe the techniques to modify the shuffling attacks regarding relaxing and data accessibility.

\par \noindent\textbf{Relaxing.}
We consider Mixing and Swapping algorithms as special cases and relatively relaxed versions of the Dominance algorithm.
Although relaxed, the attacks are still detrimental to certain data instances or groups if bypass the detection.
In practice, all three attacks can be further relaxed via certain restricting techniques including 
restricting the attack's frequency, count, region to  
decrease the shuffling attack's total occurrence for a given input.

For example,
shuffling is only successful half the time, 
shuffling only happens a maximum amount of times, 
or shuffling is more active in the high-score region than the low-score region.
If we remove line 4 in Algorithm~\ref{alg:swap}, the \textit{female} candidate in such case will be restricted to swapped only one time.

\par \noindent \textbf{Data accessibility.}
Shuffling attacks do not require data accessibility other than the protected feature, but if the adversary can access the whole training data $\mathbf{X}_\mathcal{D}$ of the model, they can trigger attacks on in-distribution data and not on out-of-distribution data. 

For example, \cite{slack2020fooling} propose a ``scaffolding'' technique that 
classifies input data as in or out of distribution.
The success of the attack depends on effectively training the ``scaffolding'' classifier.
And the attack will relax if input data deviates from the training data.
Or if the adversary can only access partial training data,
effective scaffolding may not be achievable.
However, given the same data accessibility, shuffling attacks are still achievable.
For example, the adversary can model the correlation between $\mathbf{X}_{\mathcal{S}}$ and $\mathbf{X}_\mathcal{P}$ to overcome certain cases that protected features are inaccessible from data input.
In deployed scenarios, the difference between shuffling attack and scaffolding is that the former is persistent and the latter may become non-existent after being exposed to new data.

\section{Estimating SHAP's detection capability}

In this section, we demonstrate that estimation algorithms of Shapley values may detect non-zero attributions of shuffling attacks. 
We consider a crude estimation scenario of Shapley values.

First, we assume the features are independent.
Second, we assume the features constructed a linear model.
With those two assumptions, we can estimate the Shapley values with a special case of SHAP, linear SHAP.
The SHAP values can be derived directly from the model's weight coefficients~\cite{lundberg2017unified}.

\par \noindent \textbf{SHAP detection in linear model}
We prove that, for linear models,  we can calculate
the attribution of the shuffling features leveraging the additive constraint of Shapley values. 

This allows us to estimate the effectiveness of the attacks without running the default time-consuming SHAP algorithms.
Since the shuffling attack is constructed on a group of $N$ input $x$. We use the matrix form of the formulas. For scoring function
\begin{equation}
    f(X) = \sum_{j} \beta_j X_j + 0 \cdot X_p
    \label{eqn: linearModel_single}
\end{equation}
the contribution matrix $\Phi$ given the input $X$ are:
\begin{equation}
    \Phi_j(f, X) = \beta_jX_j - \beta_j E[X_j], \ \ 
    \Phi_p(f, X) = \textbf{0}
    \label{eqn: linearSHAP_single}
\end{equation}
Due to the additive constraint of Shapley values, instance-wise feature contributions add up to the difference between the model output and the average model output,
\begin{equation}
   \Phi_p(f, X) = \textbf{0} = f(X) - E[f(X)] -\sum_j \Phi_j(f,X) 
    \label{eqn: linearSHAP_protected_f}
\end{equation}

We apply the shuffling function $h_{X_p}$ on the output vector of $f(X)$ to obtain $f'(X)$. $h_{X_p}(\cdot)$ can be omitted if all values in $X_p$ is the same:
\begin{equation}
    f'(X) = h_{X_p}(f(X)) = h_{X_p}(\sum_{j} \beta_j X_j)
\end{equation}
Since the features are assumed independent, to obtain  feature $j$'s contribution,  we set zero for all other features in $X$ including $X_p$ except $X_j$, In such condition, no shuffling occurs~(i.e., $h_{X_p}$ does not modify the output of $f$) since the values in $X_p$ are all the same~(zeros), and function $f'$ is equivalent to function $f$, 
The feature $j$'s contribution vector $\Phi_j$ follows:
\begin{equation}
    \Phi_j(f', X) = \Phi_j(f, X) 
    \label{eqn: linearSHAP_matrix}
\end{equation}

Due to the same additive constraint of Shapley values, the attribution of the protected feature follows
\begin{equation}
    \Phi_p(f', X) = f'(X) - E[f'(X)] -\sum_j \Phi_j(f',X)
     \label{eqn: linearSHAP_protected_g}
\end{equation}
$E[f'(X)]$ equals to $E[f(X)]$ since shuffling  $f$' output does not alter summary statistics such as mean or deviation.
Subtracting equation~\ref{eqn: linearSHAP_protected_g} and equation~\ref{eqn: linearSHAP_protected_f}, we get:
\begin{equation}
    \Phi_p(f', X) =  f'(X) - f(X)
    \label{eqn:shuffle_contribution}
\end{equation}

The mean of absolute attribution for protected feature $p$ is
\begin{equation}
    (1/N) \textbf{1}^T |\Phi_p| < \text{max}(|f'(X) - f(X)|)
\end{equation}
If the difference between $f'$'s outputs is small compared to the output value themselves, the unfair feature attribution may seem negligible.  

Thus, we discover the SHAP value of the protected feature in the linear rough estimation is bounded by equation~\ref{eqn:shuffle_contribution}. 
For non-linear models, equation~\ref{eqn:shuffle_contribution} can be used to obtain the shuffling feature attribution. The individual scoring feature's attribution may be difficult to calculate directly.

Other SHAP algorithms, due to various implementations that are actively being developed, may detect slightly higher or lower attributions of protected features.

\begin{figure}
    \includegraphics[width=\columnwidth]{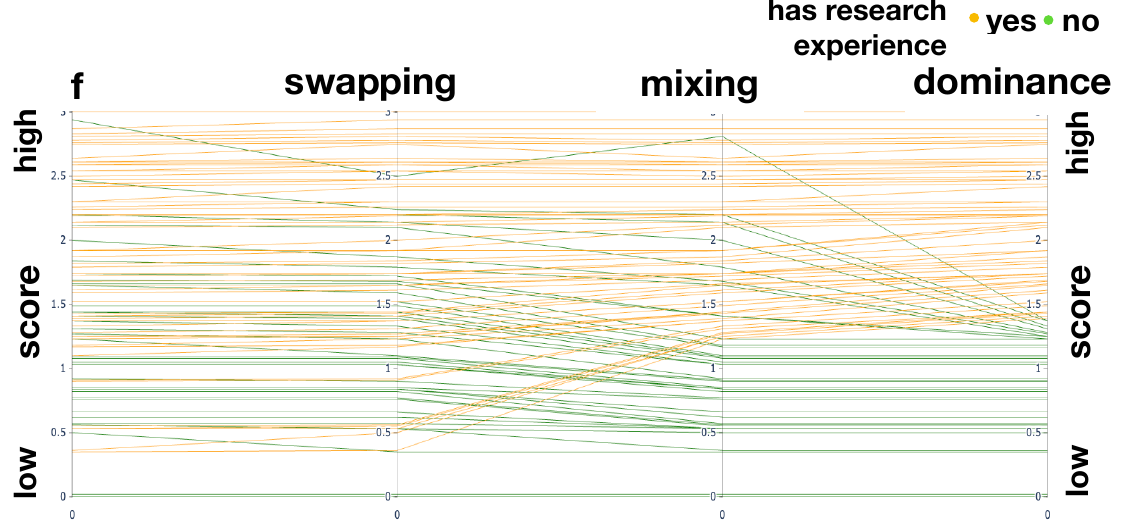}
  \caption{\textbf{Attack effect on ground truth data for admission prediction.}
  The yellow and green lines respectively indicate students with and without research experience. The yellow group's scores all drooped from the base function $f$ under different attacks (Swapping, Mixing, Dominance).}
  \label{fig:result_admission_slopPlot}
  \end{figure}
  
  \begin{figure*}
        \includegraphics[width=\textwidth]{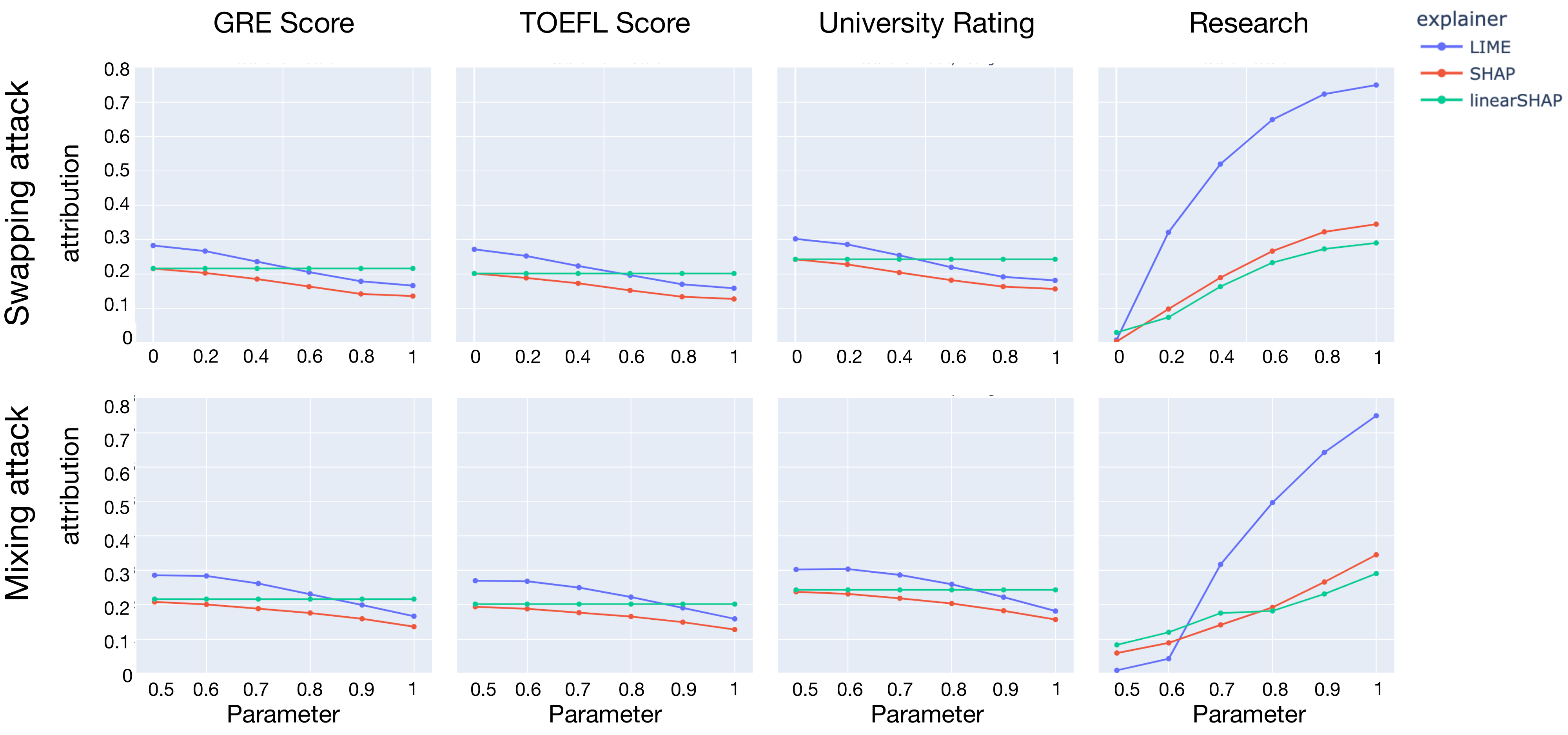}
  
      \caption{\textbf{Admission prediction experiment}
      For the three scoring features~(GRE, TOEFL, University Rating), the detected attributions from three explainers~(LIME, SHAP, and linearSHAP)  are similar. 
      For the protected feature~(Research), LIME detects more attributions as the attacks' parameters increase. A higher parameter indicates more distortion of the score vector output from the original function.
      For Swapping attack, the parameter is the percentile in the score vector that swapping is conducted multiple times to keep students with Research above students without.
      For Mixing attack, the parameter is the probability of bias towards students with Research.
      }
      
      \label{fig:result_admission}
      \end{figure*}

\section{Experiments}
We analyze the capability of SHAP to detect the attacks on three real-world datasets.\textbf{ Our data and code can be found in this anonymized link:}
https://t.ly/UkIlu.

In these experiments, first, we compare linearSHAP and SHAP~(the default option in the official SHAP python package) on attacks under different hyper-parameters.
Then, we test combinations of shuffling attacks on multiple features. 
Finally, we demonstrate how to fool SHAP using the Dominance attack and achieve a similar effect in the prior work~\cite{slack2020fooling}~ by localizing the attack.

\subsection{Datasets}
We describe here the real datasets used in the experiments, varying across admission, medical prediction, and finance.
 \par \noindent \textbf{Graduate Admission}
 Data~\cite{acharya2019comparison} includes 500 college applicant records.

 We consider three scoring features \textit{GRE}, \textit{TOEFL}, \textit{University rating}, and one protected feature \textit{Research}. 
 \par \noindent \textbf{Diabetes Risk}
Data~\cite{islam2020likelihood} includes 520 patient records with features such as \textit{Weakness}, \textit{Itching}, \textit{Irritability}. We consider all features as scoring features, except the two protected features, \textit{Sex} and \textit{Age}.
 \par \noindent\textbf{German Credit}
Data~\cite{misc_statlog_(german_credit_data)_144} includes 1000 loan applicant records including financial and demographic information. We consider one scoring feature \textit{Loan Rate\%Income} and one protected feature \textit{Gender}.

\subsection{Graduate admissions prediction}
\label{sec:admission}
We use the \textbf{Graduate admission} dataset to demonstrate the characteristics of the three attacks. The base model $f$ is equal weighted summation between X1~(GRE) X2~(TOEFL), X3~(University rating). We normalize features' data and then feed them into function $f$ or the adversarial function $f'$.
We also feed the Xp~(Research) into the functions. But the $f$ ignores it, while $f'$ uses it to manipulate the outputted score vector.

Figure~\ref{fig:result_admission_slopPlot} shows the attack effect on the ground truth data. 
The yellow and green lines indicate students with and without research experience.
It shows that the Swapping attack results in green lines downward~(or horizontal), and yellow upward~(or horizontal), indicating a clear bias that promotes the yellow group's score. 
The Mixing attack~(with probability bias of 0.8) drops more green group candidates in the lower score region than the Swapping attack, but not as much in the higher score region.
The Dominance attack results in all yellow group candidates above green group candidates.

To understand the characteristics of different attacks, we conduct experiments on the attacks under different hyper-parameters.
The attacks are conducted on Research feature.

Figure~\ref{fig:result_admission} shows the attribution detected by different explainers for all the features.
Although our attacks are designed for Shapley values or Shapley estimators~(e.g., SHAP, linearSHAP), the attacks are explainer-agnostic, and we add the LIME explainer for comparison.
Each dot on the plot is the mean absolute attribution from the 100 students' explanation outcome.

Figure~\ref{fig:result_admission}(first row) is the attribution detected from repeated Swapping attacks using the starting at different quantiles of the input data. 
A quantile value indicates the Swapping attack starts at the corresponding list position and repeats at every list position until the top.
Quantile 0 indicates attacking one time on the entire input list of scores, the default Swapping attack. 
Quantile 1 is equivalent to a Dominance attack.

Figure~\ref{fig:result_admission}(second row) is the attribution detected under Mixing attack and different bias probability increase from 0.5 to 1. 
A probability of 0.5 indicates random mixing with no bias.
Probability 1 results in a Dominance attack.

Overall, the explainers detected a similar amount of attribution from GRE, TOEFL, and University Rating, while the attribution from Research increases as the hyper-parameters increase.
Higher values of hyper-parameters indicate more distortions in the original function's outcome vector, which makes them more detectable by explainers.
However, explainers' detectability varies.
LIME detected more attributions on larger attack hyper-parameters than the SHAP and linearSHAP.
That means Shapley estimators inherited the limitation of detecting the shuffling attacks of the Shapley values. 

For SHAP and linearSHAP, the Research’s attributions are the smallest among the 4 features if the Swapping attack has a parameter of 0.2 or lower or the mixing attack has a parameter lower than 0.7. 
For the Swapping attack with parameter 0 (i.e. attack occurs only once in the entire output score vector), SHAP and LIME detected near-zero attribution from Research, but linearSHAP detected non-zero attribution.
This means the default swapping attack can fool the two widely used explainers, LIME and SHAP.

For the Mixing attack with parameter 0.5, the mixing is performed randomly without favoring students with Research. LIME detects near-zero attribution, indicating its stable detection capability under randomness. However, SHAP and linearSHAP detect significant attribution of Research.

\subsection{Diabetes prediction}
\label{sec:diabetic}

We use \textbf{Diabetes prediction} data to demonstrate when shuffling attacks are based on complicated conditions using multiple protected features~(age, and sex). 
In this experiment, we obtain a non-linear scoring function $f$ by training a logistic regression instead of defining it. 
The target label is positive or negative for diabetic diagnosis.
We train the model on 80\% training data and 20\% testing data are used in the SHP detection experiments.

$f$ outputs the probability of diabetes. 
Then we use the protected features
to construct adversarial scoring function $f'$.
We set a probability threshold of 0.9 as a positive label for all functions' output.
We construct hybrid attacks using combinations of Dominance, Mixing, and Swapping algorithms.
For each list of scores, we break it into the top half and bottom half and apply different attacks or no attack on each half.
We conduct SHAP detection on the adversarial scoring function. 
We evaluate all the adversarial scoring functions using various group fairness metrics and measured the fairness drop between $f$ and $f'$ for the gender feature~(between male and female groups).

\begin{figure}[t]
  \includegraphics[width=\columnwidth]{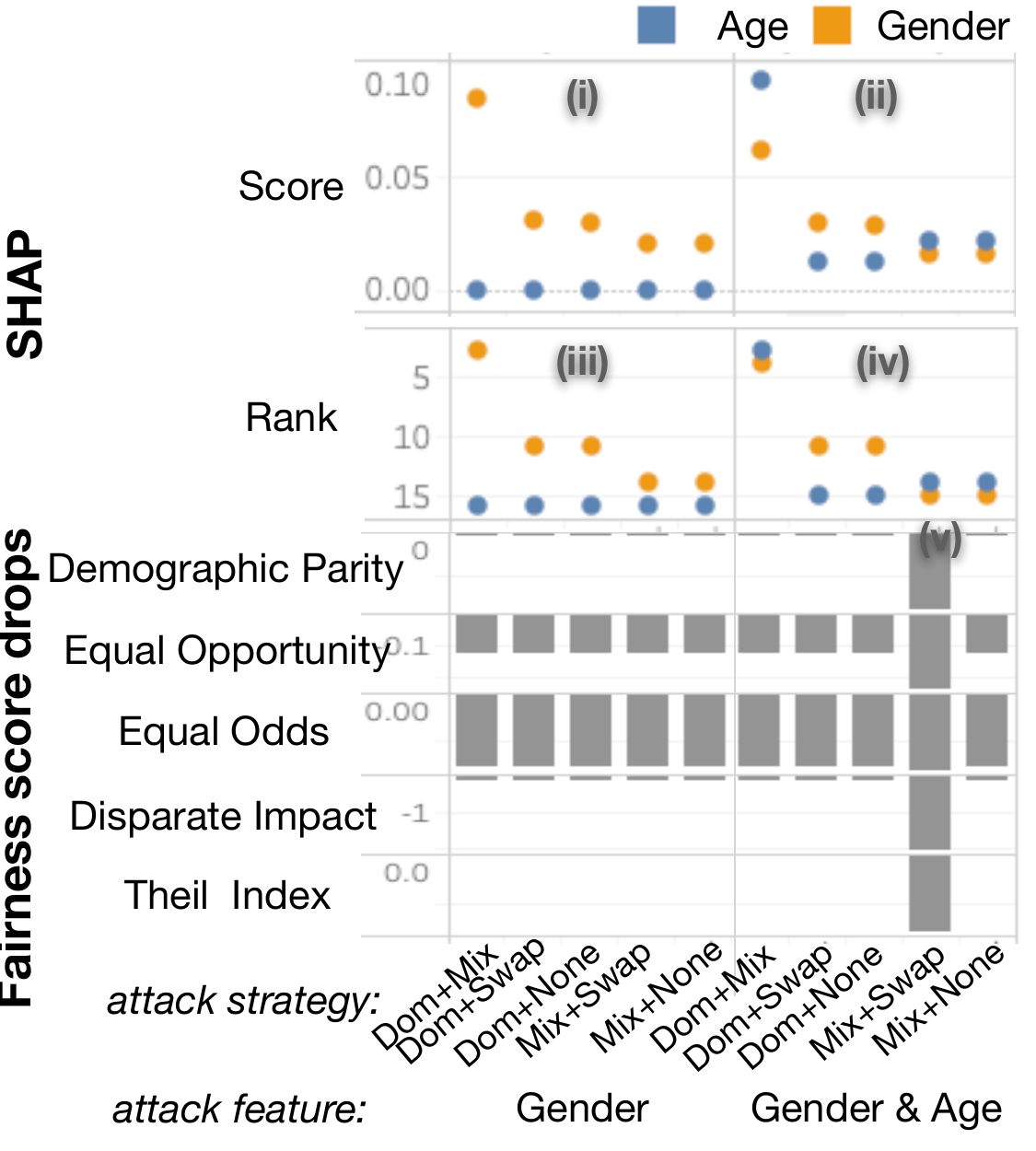}
\caption{\textbf{Diabetes prediction experiment} Each column describes the SHAP results and fairness measurements of an attack.
An attack is defined by the protected features that are used in attack construction~(gender or both gender and age), and which specific attack algorithm~(Dominance, Mixing, Swapping, or none-attack) is applied in the top and bottom half of any given input data. 
}
\label{fig:result_diabetic}
\end{figure}

Figure~\ref{fig:result_diabetic} shows in general,
When the attack uses gender~(i.e. male is the privileged group) only, the age feature's attribution is zero~(Fig.~\ref{fig:result_diabetic}i) and iii)). 
When the attack uses both gender and age~(i.e. put younger people beyond elders), the gender's attribution decreases while age's attribution increases~(Fig.~\ref{fig:result_diabetic}ii) and iv)). 
Dominance attack on the top half and Mixing attack on the bottom half~(Dom+Mix) results in the most significant feature attributions, in both rank and score~(Fig.~\ref{fig:result_diabetic}i)-iv) upper left).

The (Mix+Swap) and (Mix+None) result in the least significant feature attributions~(Fig.~\ref{fig:result_diabetic}i)-iv) lower right). 
Assuming a model auditor sets a threshold to reject models if protected features are in the top-10 ranks or higher than 0.05 in scores, only the (Dom+Mix) model is successfully detected and removed.

We also measure the group fairness drops in gender for all attack combinations.
We use the following group fairness metrics that were implemented in IBM AI Fairness360 toolkit~\cite{aif360-oct-2018}: 
\begin{itemize}
    \item \textbf{Demographic Parity}: the predicted positive rates for both groups should be the same.
    \item \textbf{Equal Opportunity}: the true positive rates for both groups should be the same. 
    \item \textbf{Equal Odds}: the true positive rates and the true negative rates for both groups should be the same.
    \item \textbf{Disparate Impact}: the ratio of positive rate for the unprivileged group to the privileged group, e) \textbf{Theil Index}: between-group unfairness based on generalized entropy indices~\cite{speicher2018unified}.
\end{itemize}

All fairness metrics scores drop from the base function $f$ to  $f'$.
The drop doesn't show a significant difference between attacks except for the (Mix+Swap) combination using both protected features~(Fig.~\ref{fig:result_diabetic}v)).

Assuming a model auditor sets fairness drop tolerance as 0.5, the (Mix+Swap) model will be removed. 
After both SHAP detection and fairness metric detection, three adversarial functions~(Dom+Swap, Dom+None, and Mix+None) attack on both features or gender alone pass the audit and may cause an unfair impact if deployed.

\subsection{Credit prediction}
\label{sec:german}

\begin{figure}[t]
  \includegraphics[width=\columnwidth]{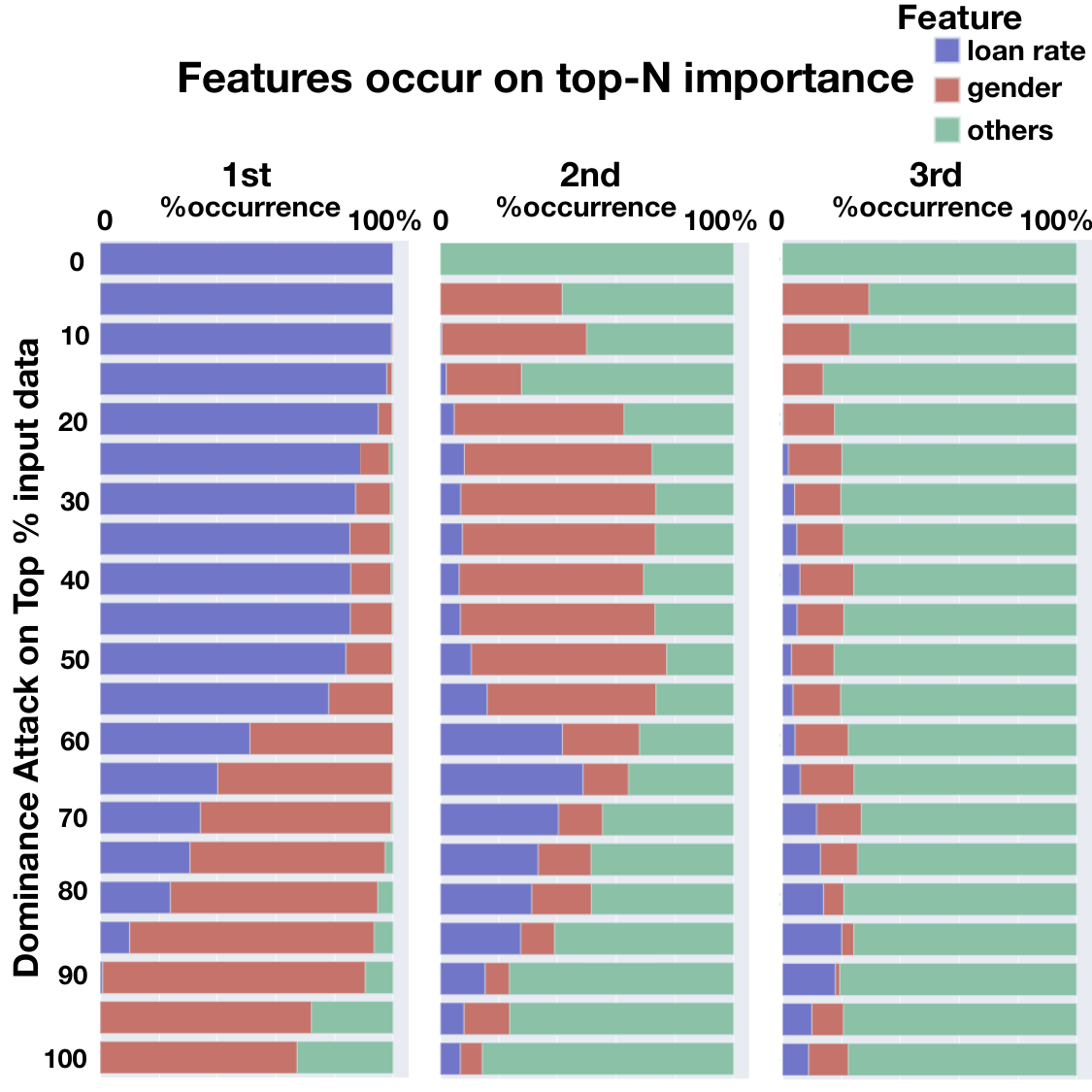}
\caption{\textbf{German credit experiment} The first row shows that if no attack happens, the loan rate occurs at top-1 importance for 100\% of the data instance. 
The last row shows that if the attack applies to all input data, the gender feature occurs at top-1 importance 67.1\%  times,
and at top-2 7.4\%.  
at top-3 13.4\%.
When the attack applies to the top 15\% of data, the gender feature has the least occurrence on top-3 importance.
}
\label{fig:result_german}
\end{figure}

Prior work demonstrates, using \textbf{German credit} data, scaffolding deception can fool SHAP.
Without such deception, the protected features~(\textit{Gender}) have the top-1 attribution among all the data instances. 
With deception, the protected features drop from top-1 to top-2 or lower importance for most data instances.

In this experiment, we aimed to achieve the same using the Dominance attack on instances at the top-$k$ percent region.
We demonstrate that with small $k$, Dominance attack can bypass SHAP detection and achieve a similar result as scaffolding deception. 
Such an attack is effective for decisions that only take top instances into account, and the effectiveness is persistent even with data shifting.

We construct the scoring function $f$ only based on \textit{Loan rate} similar to the prior work, and use \textit{Gender} feature to shuffle the output of $f$.
Fig.~\ref{fig:result_german} shows that when the attack region is the entire input data, \textit{Gender} is the 1st important feature for 67.1\% of data instances.
This encourages model auditors to investigate further. 
However, when the attack region is constrained to the top 15\% of the data instances, \textit{Gender} is only at the first importance 1.7\%,
at second importance 25.8\%, at third importance 13.7\%. 
This gives auditors less incentive for investigation.

Adversaries can engineer Dominance attacks to be more sophisticated than top-$k$ constraints.
In practice, if the Dominance attack only occurs $1$ out of $1000$ times on a target group of input, SHAP may be fooled.

\section{Discussion}

In this work, we propose three shuffling attacks that exploit the vulnerability of averaging processes in calculating Shapley values. We test two different implementations of Shapley values-based feature attribution methods, linear SHAP and SHAP, on real-world datasets to demonstrate the efficacy of the attacks. Our proposed data-agnostic attack strategies are more robust under data shifting than other data-dependent attack strategies. We theoretically prove that Shapley values are vulnerable to model output shuffling attacks.

Due to the differences in the nature of their estimation, the implementations of Shapley values, such as SHAP can detect the shuffling attack. We notice the difference between linearSHAP and SHAP in certain experimental settings. Empirically, we observed SHAP detects the dominance attack better, but linearSHAP detects the swapping attack better. The detection success varies based on the attack, underlying data, and loss function of the XAI method. 

This important finding demonstrates a counter-intuitive observation: the more accurate the Shapley value calculation algorithm is, the more vulnerable it is to the Swapping attack. For other Shapley values implementations that use different value functions, shuffling attacks can always be constructed by exploiting the process of averaging across a group of values so that the averaged outcome is invariant to shuffling.

We leverage the current XAI methods that use batch processing to obtain the vector of $n$ score outcomes from a vector of $n$ inputs. The attacks can be conducted in one function call. However, future XAI methods may call the function $n$ times to obtain a score for each item in the input vector one by one and then create one attribution score.  The attacks must be organized among multiple function calls in such cases. For example, the attack agents have to memorize previously called inputs and outputs to determine whether and how to adjust the current score.

If SHAP or other XAI methods employ consistency checks for the black-box model, the attack may be better detected and disabled. 
However, shuffling attacks can survive in many cases, such as algorithmic rankers or continuous machine learning, where consistency constraints are relaxed or removed.
Another defense against shuffling attacks is to use incomplete coalitions while calculating SHAP or use other heuristic feature attribution methods such as LIME.
We would like to emphasize that when the score difference due to Swapping is relatively small on the scale of the scores, it is still challenging. 
For example, in Figure~\ref{fig:result_admission}, LIME detects near-zero attribution from the default swapping attack~(quantile parameter is 0), and so does SHAP.

A possible defense is to have another post-processing function that scales the score differences to avoid model output cluttered in a small score range~\cite{yuan2023human}. 
But it is to be noted that such scaling is a distortion of the model behavior. 
Maintaining the usability of the explanation after distortion is part of our future work.

One limitation of the shuffling attack is its effect on the decision-maker's conclusion is not always guaranteed. For example, the conclusion will not change if the attack impacts the low-score items and the decision is based only on high-score items. Defense against the shuffling attacks can be facilitated by end-to-end communication between the developer and the user~(who may use SHAP for auditing) regarding which outcome is superior. Specifically, the algorithm~\ref{alg:template} requires such a parameter to perform the sorting in line 5.  The cost of such defense is a topic beyond the scope of this work.
We only demonstrate scenarios using protected features~(e.g. gender) directly in the shuffling attack, future work may explore using proxy features~(e.g. pregnancy status) that are correlated with the protected features.

We only tested a few Shapley estimators under the attacks~(i.e. SHAP and linearSHAP), we may explore other improved Shapley estimators such as kernelSHAP~\cite{covert2020improving} and bayeSHAP~\cite{slack2021reliable} in future work.
A recent work proves that the Generalized Linear Model's parameters can be fully recovered with $n$-Shapley value~\cite{bordt2023shapley}. 
It may be fruitful to investigate the interplay between higher-order Shapley values and the shuffling attacks.

\section{Conclusion and Future Work}

Our work demonstrated that the shuffling attack is powerful enough to fool SHAP. It is still an open problem in XAI research to develop robust detection methods for more complicated shuffling attacks.
For example, we only considered swapping between two nearby items. Still, it can be generalized to swapping between nearby items for which the behaviors are not explored. 
Generalization or derivatives of the shuffling attacks are easy to define by an attacker such as a model distributor or model broker without the need for model training. 
This relaxes the requirements of AI expertise in performing such attacks and increases the risks.
We have not yet explored the cascading impact of our unfair scoring function.  In the real world, it is common that a higher-scored candidate may receive additional advantages (e.g., getting the job offer) which leads to advantages in future scoring~(e.g., approval of a bank loan).  A small shuffling initially may result in huge future differences.
Our work opens up an alternative way of explaining model unfair behaviors: \textit{one group's score is shuffled against another in a specific way}. In the future, we will design XAI methods to generate such an explanation. 
We will develop a testing framework and user interface for interactively and collaboratively conducting attacks on current and future XAI methods using synthetic and real datasets to elicit the domain decision-makers' desiderata of decision-aid XAI systems, such as the stability and sensitivity of the XAI interpretation and decision outcome under attacks.
We will consider shuffling scores to promote candidates from non-privileged groups and investigate these fairness-preserving interventions in conjunction with established ranking fairness metrics.

\section*{Ethical Statement}

Our work aims to raise awareness about the risk of using Shapley value-based feature attribution explanations in auditing black box machine learning models. 
Through demonstrative experiments, we expose the potential negative societal impacts of relying on such explanations. 

Given the nature of our proposed shuffling attacks, where the attacks do not require access to the underlying data distribution, 
We especially want to draw the community's attention to the severity of such attacks and propose a call to action to collaboratively find more robust ways to deploy and leverage explainable AI methods in socio-technical settings.

We acknowledge that malicious model distributors or brokers could use this attack to mislead end users or cheat during an audit. However, we believe this paper increases the vigilance of the community and fosters the development of trustworthy explanation methods. 

Furthermore, by showing how data-agnostic unfairness rules can be incorporated after the model is trained, this work contributes to the research on auditing or certifying the fairness of AI-based decision systems.

\bibliography{REFERENCE}

\begin{thebibliography}{21}
\providecommand{\natexlab}[1]{#1}

\bibitem[{Acharya, Armaan, and Antony(2019)}]{acharya2019comparison}
Acharya, M.~S.; Armaan, A.; and Antony, A.~S. 2019.
\newblock A comparison of regression models for prediction of graduate
  admissions.
\newblock In \emph{2019 international conference on computational intelligence
  in data science (ICCIDS)}, 1--5. IEEE.

\bibitem[{A{\"\i}vodji et~al.(2022)A{\"\i}vodji, Hara, Marchand, Khomh
  et~al.}]{aivodji2022fooling}
A{\"\i}vodji, U.; Hara, S.; Marchand, M.; Khomh, F.; et~al. 2022.
\newblock Fooling SHAP with Stealthily Biased Sampling.
\newblock In \emph{The Eleventh International Conference on Learning
  Representations}.

\bibitem[{Arrieta et~al.(2020)Arrieta, D{\'\i}az-Rodr{\'\i}guez, Del~Ser,
  Bennetot, Tabik, Barbado, Garc{\'\i}a, Gil-L{\'o}pez, Molina, Benjamins
  et~al.}]{arrieta2020explainable}
Arrieta, A.~B.; D{\'\i}az-Rodr{\'\i}guez, N.; Del~Ser, J.; Bennetot, A.; Tabik,
  S.; Barbado, A.; Garc{\'\i}a, S.; Gil-L{\'o}pez, S.; Molina, D.; Benjamins,
  R.; et~al. 2020.
\newblock Explainable Artificial Intelligence (XAI): Concepts, taxonomies,
  opportunities and challenges toward responsible AI.
\newblock \emph{Information fusion}, 58: 82--115.

\bibitem[{Baniecki and Biecek(2023)}]{baniecki2023adversarial}
Baniecki, H.; and Biecek, P. 2023.
\newblock Adversarial Attacks and Defenses in Explainable Artificial
  Intelligence: A Survey.
\newblock \emph{arXiv preprint arXiv:2306.06123}.

\bibitem[{Bellamy et~al.(2018)Bellamy, Dey, Hind, Hoffman, Houde, Kannan,
  Lohia, Martino, Mehta, Mojsilovic, Nagar, Ramamurthy, Richards, Saha,
  Sattigeri, Singh, Varshney, and Zhang}]{aif360-oct-2018}
Bellamy, R. K.~E.; Dey, K.; Hind, M.; Hoffman, S.~C.; Houde, S.; Kannan, K.;
  Lohia, P.; Martino, J.; Mehta, S.; Mojsilovic, A.; Nagar, S.; Ramamurthy,
  K.~N.; Richards, J.; Saha, D.; Sattigeri, P.; Singh, M.; Varshney, K.~R.; and
  Zhang, Y. 2018.
\newblock {AI Fairness} 360: An Extensible Toolkit for Detecting,
  Understanding, and Mitigating Unwanted Algorithmic Bias.

\bibitem[{Bordt and von Luxburg(2023)}]{bordt2023shapley}
Bordt, S.; and von Luxburg, U. 2023.
\newblock From Shapley values to generalized additive models and back.
\newblock In \emph{International Conference on Artificial Intelligence and
  Statistics}, 709--745. PMLR.

\bibitem[{Covert and Lee(2020)}]{covert2020improving}
Covert, I.; and Lee, S.-I. 2020.
\newblock Improving kernelshap: Practical shapley value estimation via linear
  regression.
\newblock \emph{arXiv preprint arXiv:2012.01536}.

\bibitem[{Das and Rad(2020)}]{das2020opportunities}
Das, A.; and Rad, P. 2020.
\newblock Opportunities and challenges in explainable artificial intelligence
  (xai): A survey.
\newblock \emph{arXiv preprint arXiv:2006.11371}.

\bibitem[{Dimanov et~al.(2020)Dimanov, Bhatt, Jamnik, and
  Weller}]{dimanov2020you}
Dimanov, B.; Bhatt, U.; Jamnik, M.; and Weller, A. 2020.
\newblock You shouldn’t trust me: Learning models which conceal unfairness
  from multiple explanation methods.

\bibitem[{Hofmann(1994)}]{misc_statlog_(german_credit_data)_144}
Hofmann, H. 1994.
\newblock {Statlog (German Credit Data)}.
\newblock UCI Machine Learning Repository.
\newblock {DOI}: https://doi.org/10.24432/C5NC77.

\bibitem[{Islam et~al.(2020)Islam, Ferdousi, Rahman, and
  Bushra}]{islam2020likelihood}
Islam, M.; Ferdousi, R.; Rahman, S.; and Bushra, H.~Y. 2020.
\newblock Likelihood prediction of diabetes at early stage using data mining
  techniques.
\newblock In \emph{Computer Vision and Machine Intelligence in Medical Image
  Analysis}, 113--125. Springer.

\bibitem[{Laberge et~al.(2023)Laberge, A{\"\i}vodji, Hara, Marchand, and
  Khomh}]{laberge2023fool}
Laberge, G.; A{\"\i}vodji, U.; Hara, S.; Marchand, M.; and Khomh, F. 2023.
\newblock Fool SHAP with Stealthily Biased Sampling.
\newblock In \emph{International Conference on Learning Representations
  (ICLR)}.

\bibitem[{Lundberg and Lee(2017)}]{lundberg2017unified}
Lundberg, S.~M.; and Lee, S.-I. 2017.
\newblock A unified approach to interpreting model predictions.
\newblock \emph{Advances in neural information processing systems}, 30.

\bibitem[{Noppel, Peter, and Wressnegger(2023)}]{noppel2023disguising}
Noppel, M.; Peter, L.; and Wressnegger, C. 2023.
\newblock Disguising attacks with explanation-aware backdoors.
\newblock In \emph{2023 IEEE Symposium on Security and Privacy (SP)}, 664--681.
  IEEE.

\bibitem[{Ribeiro, Singh, and Guestrin(2016)}]{ribeiro2016should}
Ribeiro, M.~T.; Singh, S.; and Guestrin, C. 2016.
\newblock " Why should i trust you?" Explaining the predictions of any
  classifier.
\newblock In \emph{Proceedings of the 22nd ACM SIGKDD international conference
  on knowledge discovery and data mining}, 1135--1144.

\bibitem[{Slack et~al.(2021)Slack, Hilgard, Singh, and
  Lakkaraju}]{slack2021reliable}
Slack, D.; Hilgard, A.; Singh, S.; and Lakkaraju, H. 2021.
\newblock Reliable post hoc explanations: Modeling uncertainty in
  explainability.
\newblock \emph{Advances in neural information processing systems}, 34:
  9391--9404.

\bibitem[{Slack et~al.(2020)Slack, Hilgard, Jia, Singh, and
  Lakkaraju}]{slack2020fooling}
Slack, D.; Hilgard, S.; Jia, E.; Singh, S.; and Lakkaraju, H. 2020.
\newblock Fooling lime and shap: Adversarial attacks on post hoc explanation
  methods.
\newblock In \emph{Proceedings of the AAAI/ACM Conference on AI, Ethics, and
  Society}, 180--186.

\bibitem[{Speicher et~al.(2018)Speicher, Heidari, Grgic-Hlaca, Gummadi, Singla,
  Weller, and Zafar}]{speicher2018unified}
Speicher, T.; Heidari, H.; Grgic-Hlaca, N.; Gummadi, K.~P.; Singla, A.; Weller,
  A.; and Zafar, M.~B. 2018.
\newblock A unified approach to quantifying algorithmic unfairness: Measuring
  individual \&group unfairness via inequality indices.
\newblock In \emph{Proceedings of the 24th ACM SIGKDD international conference
  on knowledge discovery \& data mining}, 2239--2248.

\bibitem[{Yang and Stoyanovich(2017)}]{yangMeasuringFairnessRanked2017}
Yang, K.; and Stoyanovich, J. 2017.
\newblock Measuring {{Fairness}} in {{Ranked Outputs}}.
\newblock In \emph{Proceedings of the 29th {{International Conference}} on
  {{Scientific}} and {{Statistical Database Management}}}, 1--6. {Chicago IL
  USA}: {ACM}.
\newblock ISBN 978-1-4503-5282-6.

\bibitem[{Yuan et~al.(2023)Yuan, Bhattacharjee, Islam, and
  Dasgupta}]{yuan2023trivea}
Yuan, J.; Bhattacharjee, K.; Islam, A.~Z.; and Dasgupta, A. 2023.
\newblock TRIVEA: transparent ranking interpretation using visual explanation
  of black-box algorithmic rankers.
\newblock \emph{The Visual Computer}, 1--17.

\bibitem[{Yuan and Dasgupta(2023)}]{yuan2023human}
Yuan, J.; and Dasgupta, A. 2023.
\newblock A Human-in-the-loop Workflow for Multi-Factorial Sensitivity Analysis
  of Algorithmic Rankers.
\newblock In \emph{Proceedings of the Workshop on Human-In-the-Loop Data
  Analytics}, 1--5.

\end{thebibliography}

\section{Supplementary materials}

Please find all our data and code using this anonymized link:
https://t.ly/UkIlu
\end{document}